\pdfoutput=1

\documentclass[11pt]{article}

\usepackage[]{coling}

\usepackage{times}
\usepackage{latexsym}

\usepackage[T1]{fontenc}

\usepackage[utf8]{inputenc}

\usepackage{microtype}

\usepackage{inconsolata}

\usepackage{pgfplotstable}
\usepackage{tikz}
\usepackage{pgfplots}
\usetikzlibrary{positioning,matrix,calc,arrows.meta,decorations.pathreplacing,shapes.geometric}
\usepackage{float}
\usepackage{subfigure}
\usepackage{caption}
\usepackage{soul}
\usepackage{xcolor}

\usepackage{tcolorbox}
\usepackage{setspace}

\usepackage{multirow}
\usepackage{booktabs}
\usepackage{array}
\usepackage{colortbl}
\usepackage{amsmath}
\usepackage{url}
\usepackage{amssymb}

\newcommand\scalemath[2]{\scalebox{#1}{\mbox{\ensuremath{\displaystyle #2}}}}

\title{LRHP: Learning Representations for Human Preferences \\ via Preference Pairs}

\author{
    Chenglong Wang\textsuperscript{\rm 1}, 
    Yang Gan\textsuperscript{\rm 1}, 
    Yifu Huo\textsuperscript{\rm 1}, 
    Yongyu Mu\textsuperscript{\rm 1}, 
    Qiaozhi He\textsuperscript{\rm 1},
    Murun Yang\textsuperscript{\rm 1}, \\
    \textbf{Tong Xiao\textsuperscript{\rm 1,2}\footnote{Corresponding author.}, 
    Chunliang Zhang\textsuperscript{\rm 1,2}, 
    Tongran Liu\textsuperscript{\rm 3}
    and
    Jingbo Zhu\textsuperscript{\rm 1,2}} \\
    \textsuperscript{\rm 1} School of Computer Science and Engineering, Northeastern University, Shenyang, China \\
    \textsuperscript{\rm 2} NiuTrans Research, Shenyang, China \\
    \textsuperscript{\rm 3} CAS Key Laboratory of Behavioral Science, Institute of Psychology, CAS, Beijing, China \\
    {\{clwang1119, zzhu8250\}@gmail.com},
    {\{xiaotong, zhujingbo\}@mail.neu.edu.cn}
}

\begin{document}
\maketitle
\begin{abstract}
To improve human-preference alignment training, current research has developed numerous preference datasets consisting of preference pairs labeled as “preferred” or “dispreferred”. 
These preference pairs are typically used to encode human preferences into a single numerical value through reward modeling, which acts as a reward signal during reinforcement learning from human feedback (RLHF).
However, representing these human preferences as a numerical value complicates the analysis of these preferences and restricts their broader applications other than RLHF.
In contrast, in this work, we introduce a preference representation learning task that aims to construct a richer and more structured representation of human preferences.
We further develop a more generalizable framework, \textbf{L}earning \textbf{R}epresentations for \textbf{H}uman \textbf{P}references via preference pairs (namely LRHP), which extends beyond traditional reward modeling to tackle this task.
We verify the utility of preference representations in two downstream tasks: preference data selection and preference margin prediction.
Building upon the human preferences in representations, we achieve strong performance in both tasks, significantly outperforming baselines. 


\end{abstract}

\section{Introduction}

Human-preference alignment training is essential for aligning the behaviors of large language models (LLMs) with human preferences, such as reducing harmful outputs \cite{ouyang2022training,bai2022training,wang2024hybrid}.
The performance of the alignment training highly depends on rich, high-quality preference data.
Recent efforts in human-preference alignment have therefore prioritized enhancing performance by developing improved preference datasets, which typically consist of preference pairs labeled as “preferred” or “dispreferred” \cite{stiennon2020learning,pmlr-v162-ethayarajh22a,cui2023ultrafeedback,lee2023rlaif,dubois2024alpacafarm}.

Traditional alignment approaches typically use preference pairs from such datasets to encode human preferences into a single numerical value through reward modeling, which serves as a reward signal to guide model alignment during RLHF. 
However, in real-world alignment scenarios, this numerical representation of human preferences presents two key limitations:
(1) it provides an implicit form of preference learning, making it challenging to analyze specific preference features or assess the effectiveness of the learning process.
(2) it restricts the use of human preferences solely to RLHF, limiting their broader applicability.

To address these limitations, we propose a preference representation learning task. 
This task aims to learn a richer and more structured representation of human preferences rather than encoding them into a single numerical value (see Section \ref{sec:task_formulation} for its task definition). 
Our solution is to develop a more generalizable framework, \textbf{L}earning \textbf{R}epresentations for \textbf{H}uman \textbf{P}references via preference pairs (namely LRHP), which extends beyond traditional reward modeling for human preferences.
The basic idea is to encode human preferences from preference pairs into a unified representational space, enabling the capture of rich human preferences in a representation.
This representation can be used as an input for more flexible and scalable downstream applications, such as human preference analysis, preference data selection, adaptive learning strategies, etc.
Specifically, inspired by the success of sentence encoder pre-training \cite{devlin2018bert}, we introduce \texttt{<|PREFERENCE|>} as a special symbol to capture human preferences from the input preference pair, which is trained through a preference classification task.
Furthermore, we verify the utility of these representations in two downstream tasks: \textit{preference data selection} (PDS) and \textit{preference margin prediction} (PMP).

We evaluate the proposed LRHP with experiments on the fusion of nine preference datasets, encompassing over 849k preference pairs. 
Our experimental results show that: 
(1) Utilizing these representations, we effectively solve the PDS task, which improves the reusability of existing preference datasets. 
For example, using preference representation-based PDS, we can fully utilize available preference datasets to improve a reward model that targets helpfulness and harmlessness preferences, achieving an improvement of 4.57 points on helpfulness accuracy.
(2) We develop a predictor that takes preference representations and generates preference margin scores for the PMP task.
This predictor can be well-trained with a few labeled samples.
When aligning LLMs with DPO, the predicted margin score can be a superior alternative to a rule-based margin, such as length ratio, acting as constraints to improve performance \cite{meng2024simpo,zhou2024prior}.
(3) Specific preference features can be effectively analyzed through visualizing representations learned by LRHP.


\section{Related Work}
\paragraph{Human-Preference Alignment.}
RLHF has effectively aligned LLM behaviors with human preferences \cite{stiennon2020learning,ouyang2022training}. 
Several works have improved RLHF by using fine-grained reward models \cite{wu2024fine}, reward model ensembles \cite{coste2023reward}, and refined optimization process \cite{wang2024esrl,wang2024hybrid}.
To avoid the need to train a reward model for RLHF, \citet{rafailov2024direct} proposed direct preference optimization (DPO).
Building on this approach, further research explored different variants \cite{hong2024reference,meng2024simpo,zhou2024prior}.
In practice, the performance of these alignment approaches is highly dependent on the quality of the human preference data.
Consequently, significant efforts have been devoted to constructing preference datasets to improve RLHF in LLMs, including task-specific \cite{stiennon2020learning,xu2024contrastive} and general preference datasets \cite{bai2022training,cui2023ultrafeedback}.
However, the alignment approaches limit the use of such datasets to encode human preferences into a numerical value, restricting their broader applications other than RLHF, which motivates us to propose the preference representation learning task.

\paragraph{Representation Learning in NLP.}
Representation learning in NLP aims to learn representations of textual data that are useful for classification and prediction \cite{liu2021representation,xiao2023introduction}.
A predominant approach is the pre-training of language models \cite{devlin2018bert,liu2019roberta,brown2020language}. 
For example, \citet{devlin2018bert} pre-trained an encoder-only language model through masked token prediction and next sentence prediction tasks, allowing the model to learn a representation of the input text within the \texttt{[CLS]} token.
To improve representation learning, recent works investigated the refinement of training objectives \cite{jiang2020robust,di2021efficient} and model architecture \cite{lewis2019bart,chung2024scaling}. 
Apart from these improvements, another strand of research in representation learning focused on designing probing tasks. 
These tasks used the learned representations to train additional classifiers, which aim to determine whether a model encapsulates the feature of the language, such as syntax and semantics, within its representation \cite{wallace2019nlp,hernandez2022ast}.
Inspired by the success of prior research, we explore preference representation learning.

\begin{figure*}[t]
    \centering
    \includegraphics[width=1.0\linewidth]{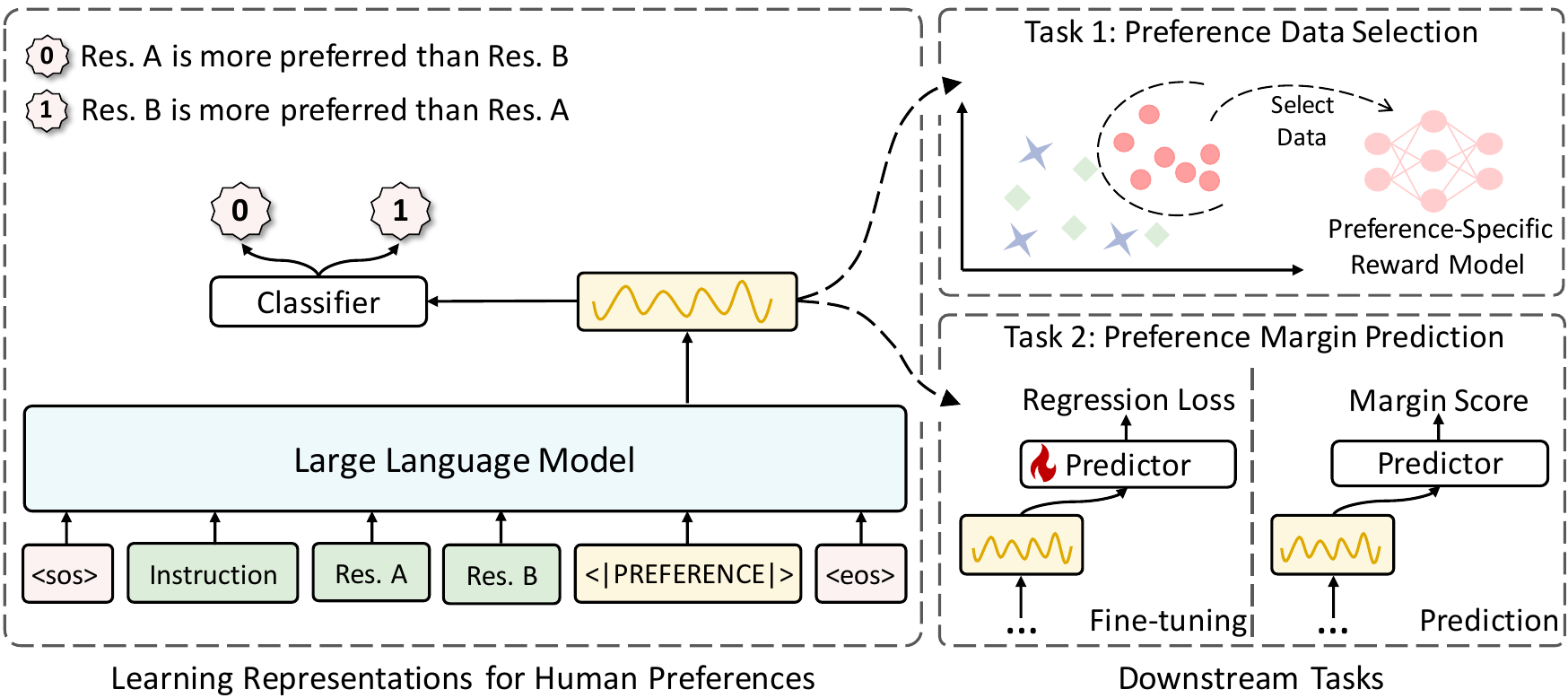}
    \vspace{-7mm}
    \caption{
    The overview of LRHP. 
    The left panel illustrates learning representations for human preferences via comparison pairs. 
    The right panel shows two downstream tasks designed to verify these preference representations.
    }
    \vspace{-5mm}
    \label{fig:main_image}
\end{figure*}

\section{Background}
\paragraph{Reinforcement Learning with Human Feedback.}
RLHF typically consists of two main steps: 1) training a reward model from preference data, and 2) using an RL algorithm like PPO \cite{schulman2017proximal}, to maximize the reward.
In step 1, we often use the Bradley-Terry model \cite{bradley1952rank}, where the preference data typically 
exists as a preference pair.
The loss function is:
\begin{equation}
    \scalemath{0.9}{
    \mathcal{L}_{reward}=-\log (\sigma (r_{\theta}(x,y_{w})-r_{\theta}(x,y_{l})))}
    \label{eq:reward_training_1}
\end{equation}
where $\sigma$ is the Sigmoid activation function, $r(\cdot)$ is a reward model and $\theta$ is its parameters.
$y_{w}$ and $y_{l}$ are two different responses for the human prompt $x$, where $y_{w}$ is more preferred than $y_{l}$.
This optimization process effectively encodes the human preferences from the pairs into numerical values.
In step 2, these values serve as signals for adjusting the parameters of the language models.

\paragraph{Direct Preference Optimization.}
To bypass the need to train a reward model, \citet{rafailov2024direct} proposed the direct preference optimization (DPO) which employs a reward model training objective to maximize rewards.
It gives a new loss function:
\begin{equation}
\scalemath{0.80}{
    \mathcal{L}_{\mathrm{DPO}}=-\log \sigma [ \beta \log(\frac{p_{\theta'}(y_{w}|x)}{p_{\theta'_{old}}(y_{w}|x)}) -\beta \log(\frac{p_{\theta'}(y_{l}|x)}{p_{\theta'_{old}}(y_{l}|x)})]
    }
\end{equation}
where $\theta'$ denotes the parameters of the language model, $\theta'_{old}$ denotes the parameters of the language model trained via supervised fine-tuning (SFT), and $\beta$ denotes a scaling factor.

To control the difference between preferred and dispreferred responses during DPO optimization, recent works introduce a constraint into the DPO objective, incorporating a preference margin $M$:
\begin{equation}
\scalemath{0.85}{
\begin{aligned}
    \mathcal{L}_{\mathrm{ConstrainedDPO}}=&-\log \sigma [ \beta \log(\frac{p_{\theta'}(y_{w}|x)}{p_{\theta'_{old}}(y_{w}|x)})\\& -\beta \log(\frac{p_{\theta'}(y_{l}|x)}{p_{\theta'_{old}}(y_{l}|x)}) - M]
\end{aligned}}
\label{eq:constrained_dpo}
\end{equation}
where $M$ is typically defined as the difference in length between responses \cite{park2024disentangling}, the probability of the preferred response \cite{xu2024contrastive}, or a fixed value \cite{meng2024simpo}.

\paragraph{Best-of-$n$ Sampling.}
Best-of-$n$ sampling refers to reordering a set of candidate responses sampled from a trained model \cite{lee2021discriminative,fernandes2022quality}.
Given a set $\mathcal{Y}$ of $n$ candidate responses for $x$, we can also use the best-of-$n$ sampling approach to align the response with human preferences.
Typically, we employ the reward model to score the candidate responses and select a response with the highest reward score.

\section{Preference Representation Learning}
This work aims to present a richer representation of human preferences.
We introduce a preference representation learning task and tackle it using our LRHP as illustrated in Figure \ref{fig:main_image}.

\subsection{Task Definition}
\label{sec:task_formulation}
The goal of preference representation learning is to learn a function that maps a preference pair to a lower-dimensional space, where the structure of this space captures the underlying human preferences.
Formally, given a preference pair $P$, the task is to learn a function $f(\cdot)$:$P\rightarrow \mathbb{R}^d$ where $f(\cdot)$ projects the preference pair into a $d$-dimensional space.
The learned function $f(\cdot)$ should ensure that $f(P_{i}) \approx f(P_{j})$ when the human preferences between preference pairs $P_{i}$ and  $P_{j}$ are highly similar, and $f(P_{i}) \gg f(P_{j})$ (or $f(P_{i}) \ll f(P_{j})$) when the preferences differ significantly.

\subsection{Learning Representations for Human Preferences via Preference Pairs}

\paragraph{Input/Output Representations.}
Starting from the SFT LLM with the final unembedding layer removed, we construct a preference representation model to achieve the function $f(\cdot)$.
The input can be an arbitrary preference pair concatenated into a single text sequence.  
Each sequence concludes with a special token \texttt{<|PREFERENCE|>}.
The final hidden state of this token serves as the representation, capturing the underlying human preferences within this input preference pair. 
A visualization of this structure is depicted in Figure \ref{fig:main_image}(left).

\paragraph{Optimization via Preference Classification.}
We posit that the labels “preferred” and “dispreferred” encapsulate rich human preferences.
To leverage this, we optimize our preference representation model using these labels, enabling the \texttt{<|PREFERENCE|>} representation to capture these features.
Specifically, we implement this optimization through a preference classification task, which assigns a label of “0” or “1” to each preference pair. 
A label of “0” indicates that the first response in the input sequence is more preferred, while a label of “1” signifies a preference for the second response.
We achieve the optimization objective using binary cross-entropy loss. 
To ensure dataset balance, we randomly shuffle the response order within each pair, maintaining an equal number of samples for each class.
Additionally, we explore alternative optimization schemes to learn preference representations (see Appendix \ref{app:more_analysis}).

\subsection{Downstream Tasks for LRHP}
The learned representations can serve as feature inputs for more flexible and scalable downstream applications. 
To verify their effectiveness, we evaluate them using two downstream tasks as instances.

\subsubsection{Task 1: Preference Data Selection}
In human-preference alignment training, training a specific reward model remains a significant challenge.
It relies on a substantial amount of preference-specific data annotated by humans \cite{dubois2024alpacafarm}.
Addressing this problem, we introduce a preference data selection (PDS) task, which minimizes the need for extensive annotations by selecting data that aligns with specific preferences from the available preference datasets.
As an example, when constructing a reward model for a specific preference, we define the PDS task as follows: 
Given an available preference dataset $\mathcal{D} =\{s_{1}, s_{2},\cdots,s_{m}\}$ and a preference-specific dataset $\mathcal{D}_{\mathrm{PS}}=\{s_{1}^{\mathrm{PS}},s_{2}^{\mathrm{PS}},\cdots,s_{n}^{\mathrm{PS}}\}$, where $m$ and $n$ denote the number of preference samples in $\mathcal{D}$ and $\mathcal{D}_{\mathrm{PS}}$, respectively.
Here, $m$ is much larger than $n$.
The PDS task aims to select data from $\mathcal{D}$ that closely matches the preferences in $\mathcal{D}_{\mathrm{PS}}$, thus reducing the need for preference-specific annotations.

Leveraging preference representations learned by LPHP, including preference types and tasks as described in Figure \ref{fig:different_layers}, we address the PDS task by matching representations.
Specifically, we define a distance score of the $i$-th sample in $\mathcal{D}$ by:
\begin{equation}
    \scalemath{0.90}{
    C_{i} = \frac{1}{n}\sum_{j}^{n}\mathrm{Match}(f(s_{i}), f(s_{j}^{\mathrm{PS}}))}
    \label{eq:distance_score}
\end{equation}
where $f(s_{i})$ and $f(s_{j}^{\mathrm{PS}})$ denote the representations of the samples $s_{i}$ and $s_{j}^{\mathrm{PS}}$, respectively.
$\mathrm{Match}(\cdot)$ is a matching function which computes the distance between representations.
A smaller distance score indicates that the preference sample more closely matches the preferences in $\mathcal{D}_{\mathrm{PS}}$.
Here, we employ the cosine function as the matching function.

In this work, we apply this approach to select preference data from available datasets to improve a reward model that targets harmlessness and helpfulness preferences, providing a practical instance.

\subsubsection{Task 2: Preference Margin Prediction}
\label{sec:task2}
In constructing a constrained DPO, the ideal preference margin should reflect the overall degree of difference, as outlined by \citet{touvron2023llama}. 
However, their approach depends on a large number of annotators to establish the preference margin, which poses a challenge for maintaining consistency across the available preference datasets.
To overcome this challenge, we present a preference margin prediction (PMP) task, where the margin for each comparison pair is predicted through fine-tuning on a small set of labeled samples.
We utilize the learned preference representations to perform the PMP task through an adaptive learning strategy.
Specifically, following \citet{touvron2023llama}'s work, we manually label preference margin scores on a four-point scale to a subset of comparison pairs from existing datasets. 
These labeled samples are then used to fine-tune our preference representation model coupled with a regressive predictor. 
The model and predictor generate the preference margin scores for the remaining comparison pairs.
The predicted margin scores are used as the $M$ in Eq. \ref{eq:constrained_dpo}.
More details on margin score labeling and model fine-tuning are provided in Appendix \ref{app:experimental_details}.

\section{Experiments}
\subsection{Experimental Setups}
We evaluated LRHP on the downstream tasks of PDS and PMP using the LLaMA-3-8B-Instruction and Mistral-7B-Instruction models.
\subsubsection{Datasets}
\paragraph{Datasets for Optimizing Representation Model.}
We trained the model using 849k preference pairs drawn from (1) four general preference datasets, including Anthropic Helpful\&Harmless \cite{bai2022training}, Alpacafarm \cite{dubois2024alpacafarm}, SHP \cite{pmlr-v162-ethayarajh22a}, and UltraFeedback \cite{cui2023ultrafeedback} and (2) five task-specific preference datasets, including WebQA \cite{nakano2021webgpt}, summarization \cite{stiennon2020learning}, math reasoning \cite{lai2024step}, code generation \cite{CodeFeedback-Filtered-Instruction}, and machine translation \cite{xu2024contrastive}.
We performed binarized preference pair processing on UltraFeedback with four responses per instruction to ensure format uniformity, following \citet{ouyang2022training}.
The detailed construction of this data can be found in Table \ref{tab:fusion_preference_pairs}.

\paragraph{Datasets for PDS Task.}
We used Anthropic Helpful\&Harmless as our preference-specific preference dataset $\mathcal{D}_{\mathrm{PS}}$.
We used all preference pairs used by optimizing the representation model as the dataset $\mathcal{D}$ that will be selected.
Note that we excluded preference pairs similar to $\mathcal{D}_{\mathrm{PS}}$ from $\mathcal{D}$.
To reduce computational overhead, we selected a representative subset from $\mathcal{D}_{\mathrm{PS}}$, which consisted of 2k helpful and 2k harmful preference pairs.

\paragraph{Datasets for PMP Task.}
We labeled the margin scores for 3.2k preference pairs from preference pairs used by optimizing the representation model.
Of these, 3k were used to fine-tune our preference representation model, and the remaining pairs were used for testing.
The fine-tuned model was then employed to predict preference margin scores.

\subsubsection{Evaluation}

\paragraph{PDS Task.}
The reward model was trained using selected data from the PDS.
The performance of the PDS was evaluated by measuring the quality of the reward model.
Specifically, we evaluated the trained reward model's \textit{preference accuracy} on test preference pairs.
Additionally, we evaluated the quality of the trained reward model when applied to human-preference alignment training through best-of-$n$ sampling and RL.
In this way, the trained reward models were evaluated by comparing their GPT-4 win rate, where the responses from the SFT model served as the baseline.

\paragraph{PMP Task.}
We evaluated the PMP task in the following two ways. 
Initially, we computed the level of correlation, including both Spearman and Pearson, between the predicted scores and human labeling.
Subsequently, we evaluated the model, which was trained using the modified UltraFeedback dataset (\textit{i.e.}, \texttt{llama3-ultrafeedback}\footnote{\url{https://huggingface.co/datasets/princeton-nlp/llama3-ultrafeedback}} and \texttt{mistral-instruct-ultrafeedback}\footnote{\url{https://huggingface.co/datasets/princeton-nlp/mistral-instruct-ultrafeedback}}) and predicted preference margin scores, through two commonly used benchmarks: AlpacaEval2 \cite{li2023alpacaeval} and Arena-Hard \cite{li2024live}.
We employed \texttt{GPT-4} as a proxy for human evaluation of response quality and reported the raw win and length-controlled win rates against the baseline model.
The length-controlled metric was specifically employed to counteract model verbosity.

\begin{figure}[t]
    \centering
    \vspace{1mm}
    \includegraphics[width=1.0\linewidth]{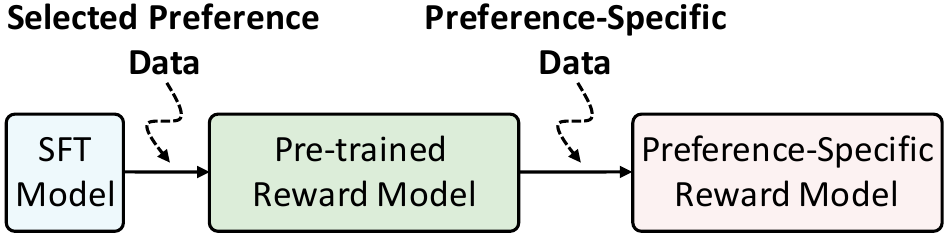}    
    \caption{
    To improve the preference-specific reward model using PDS, we first pre-train the model on selected preference data, followed by fine-tuning it with a limited amount of preference-specific data.
    This strategy allows the reward model to inherit the same preferences and reduce the need for preference-specific data.
    }
    \vspace{-5mm}
    \label{fig:incorporate_selected_data}
\end{figure}

\subsubsection{Settings}
We initialized the representation model using the LLaMA-3-8B-Instruction model. 
For the PDS task, we incorporated the selected preference data to train a preference-specific reward model in sequence as shown in Figure \ref{fig:incorporate_selected_data}.
This effectiveness was supported by the findings of \citet{kang2024get} and \citet{wang2024rovrm}.
To evaluate the performance of these reward models, we employed best-of-$n$ sampling and PPO as our basic algorithms.
We conducted our experiments using the LLaMA-3-8B-Instruction and Mistral-7B-Instruction-v0.2 models, respectively.
For the PMP, we performed preference optimization using the SimPO framework and applied the same hyper-parameters as outlined in \citet{meng2024simpo}.
More details on training are shown in Appendix \ref{app:experimental_details}.

\subsubsection{Baselines}
For the preference representation-based PDS, our baseline was the standard reward model training (\textbf{Vanilla}), \textit{i.e.}, training a preference-specific reward model using only preference-specific data.
To evaluate the effectiveness of preference representations in preference data selection, we chose \textbf{PDS-Random} as a baseline.
In PDS-Random, we randomly selected samples during the preference data selection.
In addition to PDS-Random, we also compared with other data selection approaches, including an embedding-based approach (\textbf{PDS-Emb}), where the preference representation was replaced by the \texttt{<eos>} embedding from the LLaMA-3-8B-Instruction model.


For the PMP, our baseline involved directly training a preference margin prediction model using an SFT model (\textbf{PMP-SFT}), instead of fine-tuning a preference representation model on the labeled data.
Moreover, we compared hand-designed preference margin scores, including a length difference (\textit{i.e.}, \textbf{R-DPO}) \cite{park2024disentangling}, a Kullback-Leibler (KL) constraint (\textit{i.e.}, \textbf{KTO}) \cite{ethayarajh2024kto}, and a fixed value (\textit{i.e.}, \textbf{IPO} and \textbf{SimPO}) \cite{azar2024general,meng2024simpo}.
We also chose other DPO variants as our baselines, including \textbf{RRHF} \cite{yuan2024rrhf} and \textbf{CPO} \cite{xu2024contrastive}.

\begin{table}[t]
    \centering
    \scalebox{0.75}{
    \begin{tabular}{lcccc}
\toprule[1.1pt]
\multirow{2}{*}{Reward Model} & \multicolumn{2}{c}{LLaMA-3} & \multicolumn{2}{c}{Mistral}\\  \cmidrule(l){2-3} \cmidrule(l){4-5}
& Helpful.      & Harmless.  & Helpful.      & Harmless.    \\ \midrule
Vanilla      &70.52   &71.84 &62.31   &63.32 \\ \midrule
PDS-Random   &71.23   &72.41 &61.84   &64.30 \\
PDS-Emb      &73.01   &73.53  &62.91   &65.31 \\
\rowcolor{gray!20}
PDS-LRHP     &\bf{75.09}   &\bf{75.82} &\bf{65.89}  & \bf{67.68} \\
\bottomrule[1.1pt]
\end{tabular}}
    \caption{
    Preference accuracy of the PDS task on helpfulness (Helpful.) and harmlessness (Harmless.), using the LLaMA-3-8B-Instruction and Mistral-7B-Instruction models. 
    The best results for each group are in bold.
    }
    \label{tab:preference_accuracy_pds}
    \vspace{-5mm}
\end{table}

\subsection{Experimental Results}
\subsubsection{Results of Preference Data Selection}
\paragraph{Preference Accuracy.}
Table \ref{tab:preference_accuracy_pds} summarizes the performance of our PDS in terms of preference accuracy.
Across both helpfulness and harmlessness preferences, PDS-LRHP consistently outperforms the Vanilla model which does not use PDS.
Notably, when using the LLaMA-3 model, PDS-LRHP can outperform Vanilla by 4.57 points in helpfulness accuracy.
This trend is also observed with the Mistral model.
Additionally, compared to LRHP and other PDS baselines, LRHP achieves superior preference accuracy. 
For example, when using the LLaMA-3 model to train a reward model, LRHP outperforms Emb by 2.08 points in helpfulness accuracy.
This improvement can be attributed to PDS-LRHP's use of richer preference representations, compared to approaches that rely on embeddings.

\paragraph{Best-of-$n$ Sampling\&Reinforcement Learning.}

\begin{figure}[t]
    \centering
    \includegraphics[width=1.0\linewidth]{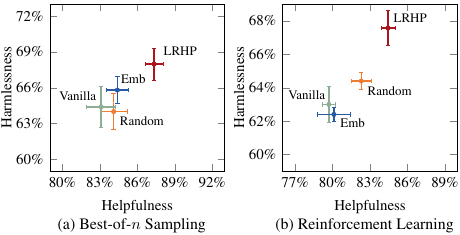}   
    \vspace{-7mm}
    \caption{
    We report the performance of the PDS task using the LLaMA-3-8B-Instruction model, with the helpfulness win rate on the x-axis and the harmlessness win rate on the y-axis.
    Experiments are conducted using three different seeds, and we report the average results along with their standard deviations.
    }
    \label{fig:pds_bos_rl_llama2}
\end{figure}

\begin{figure}[t]
    \centering
    \hspace{-1mm}
    \includegraphics[width=1.0\linewidth]{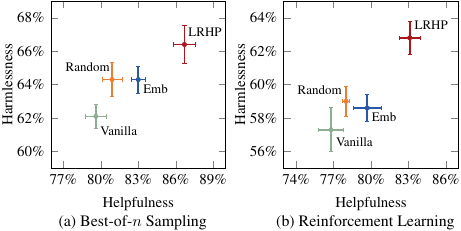}  
    \vspace{-7mm}
    \caption{
    We also report the performance of the PDS task using the Mistral-7B-Instruction model.
    }
    \vspace{-5mm}
    \label{fig:pds_bos_rl_llama3}
\end{figure}

We also evaluate the quality of the trained reward model through best-of-$n$ sampling and RL training, with the results shown in Figures \ref{fig:pds_bos_rl_llama2} and \ref{fig:pds_bos_rl_llama3}.
In best-of-$n$ sampling, LRHP consistently outperforms all baselines on the PDS task, highlighting that the human preferences captured by LRHP's representations are highly advantageous for PDS.
Unlike best-of-$n$ sampling, RL typically requires a more robust reward model, as it must not only classify responses as “good” or “bad” but also provide an accuracy-based margin between responses.
From the results, we find that PDS-LRHP more effectively meets this requirement compared to other baselines, resulting in improved RL training performance on both the LLaMA-3 and Mistral models.

Furthermore, compared to PDS-Random, PDS-LRHP demonstrates better performance across all benchmarks. 
This demonstrates that preference representation-based preference data selection surpasses random selection in effectiveness.
However, in some cases, PDS-Random also achieves a slight improvement over Vanilla. 
We attribute this improvement to PDS-Random's collection of some auxiliary preferences during the training of a preference-specific reward model.

\begin{figure}[t]
    \centering
    \vspace{1.0mm}
    \includegraphics[width=1.0\linewidth]{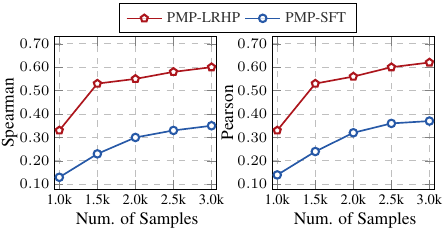}  
    \vspace{-7mm}
    \caption{
    Comparison of PMP-SFT and PMP-LRHP across different numbers of labeled samples.
    }
    \vspace{-1mm}
    \label{fig:pmp_difference_sizes}
\end{figure}

\begin{table}[t]
    \centering
    \vspace{1mm}
    \scalebox{0.68}{
    \begin{tabular}{lcccccc}
\toprule[1.1pt]
\multirow{3}{*}{Method} & \multicolumn{3}{c}{LLaMA-3} & \multicolumn{3}{c}{Mistral}\\ \cmidrule(l){2-4} \cmidrule(l){5-7} 
& \multicolumn{2}{c}{Alpaca.}     & Arena.  & \multicolumn{2}{c}{Alpaca}     & Arena.  \\ \cmidrule(l){2-3} \cmidrule(l){4-4} \cmidrule(l){5-6} \cmidrule(l){7-7}
&LC  &WR   &WR  &LC  & WR   &WR   \\ \midrule
SFT   &26.0  &25.3         &22.3    & 17.1 &14.7  &12.6          \\ \midrule
\multicolumn{5}{l}{\textbf{\textit{w/o Preference Margin}}}   \\ \midrule
DPO                 &40.3   &37.9   &32.6   &26.8    &24.9  &16.3     \\
RRHF                &31.3   &28.4   &26.5    &25.3        &24.8  &18.1\\
CPO                 &28.9   &32.2   &28.8 &23.8  &28.8 &22.6           \\ \midrule
\multicolumn{5}{l}{\textbf{\textit{w/ Preference Margin}}}   \\ \midrule
KTO (KL)            &33.1   &31.8   &26.4    &24.5        &23.6 &17.9\\     
R-DPO (Length-Diff) &41.1   &37.8   &33.1     &27.3        &24.5  &16.1\\
IPO (Fixed)         &35.6   &35.6   &30.5 &20.3 &20.3 &16.2           \\
SimPO (Fixed)       &44.7   &40.5   &33.8 &32.1 &34.8  &21.0           \\
\rowcolor{gray!20}
Ours (PMP-SFT)      &43.2     &40.7      &31.4 &32.5 &33.9 & 21.5          \\
\rowcolor{gray!20}
Ours (PMP-LRHP)     &\bf{45.6}&\bf{43.8} &\bf{35.7} &\bf{34.2} &\bf{36.7} &\bf{22.8}        \\
\bottomrule[1.1pt]
\end{tabular}}
    \caption{
    The raw win rates (WR) and length-controlled win rates (LC) for different DPO variants on AlpacaEval2 (Alpaca.) and Arena-Hard (Arena.), using the LLaMA-3-8B-Instruction and Mistral-7B-Instruction models. 
    The approach for obtaining preference distance is indicated in parentheses.
    Partial baseline results are taken from \citet{meng2024simpo}.
    }
    \vspace{-5mm}
    \label{tab:pmp_dpo}
\end{table}

\subsubsection{Results of Preference Margin Prediction}
\paragraph{Correlation Coefficient.}
We fine-tune our representation model and SFT model (\textit{i.e.}, LLaMA-3-8B-Instruction) using different numbers of labeled preference margin samples, respectively.
For the SFT model, the predictor utilizes the \texttt{<eos>} representation as input.
Figure \ref{fig:pmp_difference_sizes} depicts the correlation coefficients with human labeling.
From the results, we observe that our representation model consistently outperforms the SFT model across different sample sizes.
This observation demonstrates the significant advantage of the LRHP in predicting preference margins through its learned preference representation.
Similar to the success of fine-tuning pre-trained language models like BERT \cite{devlin2018bert}, LRHP performs well even with a limited amount of labeled data.
For example, LRHP can obtain a Spearman correlation of 0.604 with just 3k labeled samples.

\paragraph{Preference Optimization.}
We further validate the effectiveness of the predicted preference margin scores in DPO training by using these scores as the margin $M$ in Eq. \ref{eq:constrained_dpo}. 
Table \ref{tab:pmp_dpo} presents the corresponding experimental results.
First, compared to hand-designed preference margins, PMP-LRHP achieves a significant improvement, surpassing SimPO by 3.3 points in raw win rate on the AlpacaEval2 benchmark.
Beyond the raw win rate, our PMP-LRHP also effectively reduces length bias and obtains a 45.6 length-controlled win rate on the LLaMA-3 model.
We conjecture that this reduction in length bias could be attributed to our margin score's ability to provide a comprehensive evaluation. 
For example, the annotator assigns a smaller margin to longer and shorter responses with minimal meaning difference, potentially minimizing this length bias.
Furthermore, PMP-LRHP consistently outperforms DPO and its variants by a significant margin, obtaining an improvement of 6.5 points on the Arena-Hard benchmark over standard DPO when aligning the Mistral model.
We also observe that PMP-LRHP demonstrates better consistency than PMP-SFT, as seen in Figure \ref{fig:pmp_difference_sizes}.
These observations again demonstrate the effectiveness of preference representations.

\begin{figure}[t]
    \centering
    \includegraphics[width=1.0\linewidth]{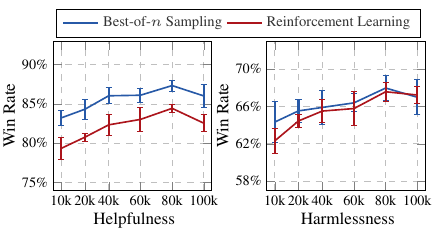} 
    \vspace{-7mm}
    \caption{
    We train preference-specific reward models with different numbers of selected preference pairs.
    }
    \vspace{-5mm}
    \label{fig:diff_sizes_error}
\end{figure}

\begin{figure*}
    \centering
    \includegraphics[width=1.0\linewidth]{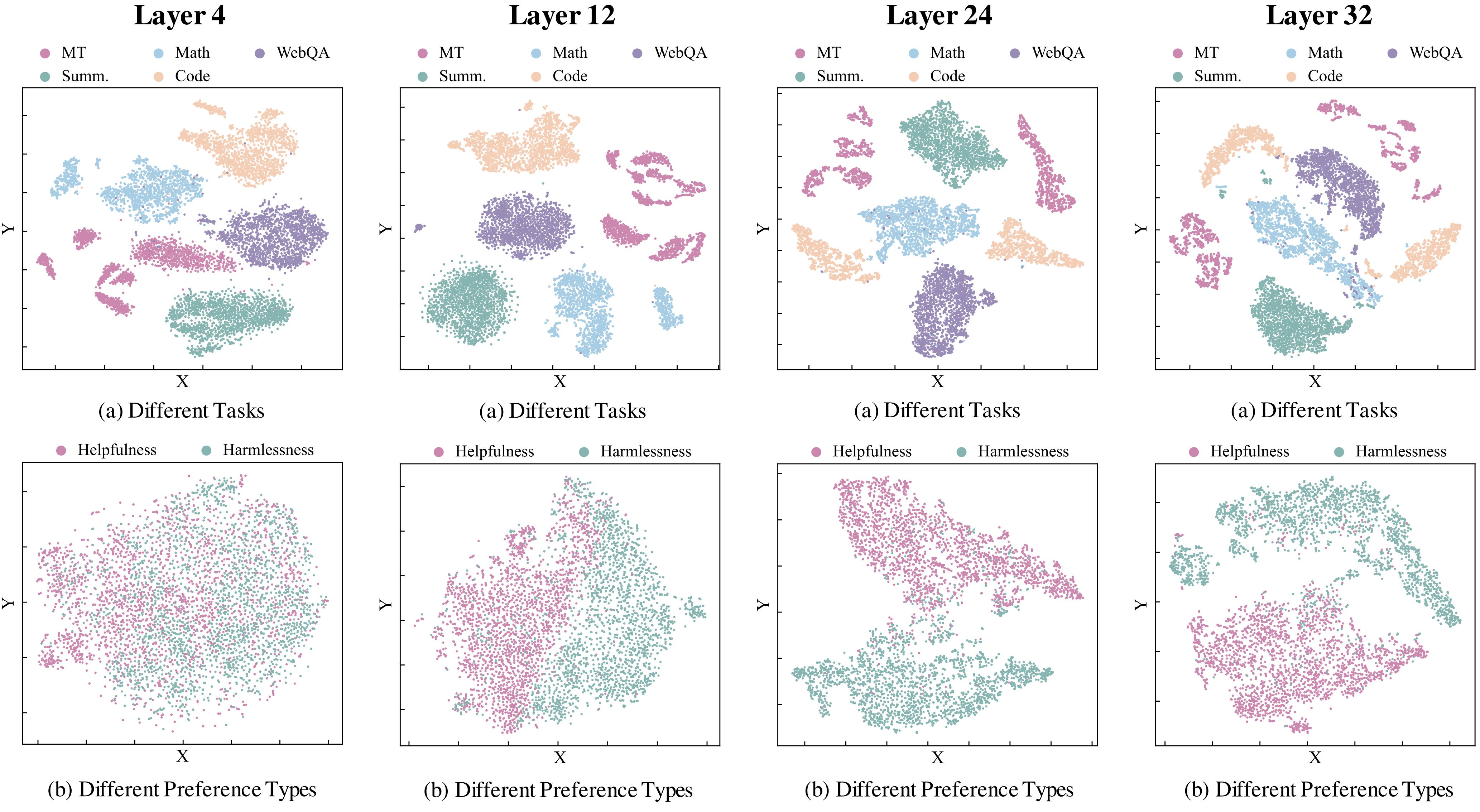} 
    \vspace{-7mm}
    \caption{
    We visualize the learned preference representations for the preference type and task at different layers.
    We apply the T-SNE algorithm, projecting the representations into two dimensions.
    }
    \vspace{-5mm}
    \label{fig:different_layers}
\end{figure*}

\subsection{Analysis}
\paragraph{Performance on Different Numbers of Selected Preference Pairs.}
We investigate the impact of different numbers of selected preference pairs in the PDS task.
We perform pre-training on a reward model using preference pairs of 10k, 20k, 40k, 60k, 80k, and 100k and then test the reward models through best-of-$n$ sampling and reinforcement learning. 
The experiments are conducted on the LLaMA-3-8B-Instruction model, with the results summarized in Figure \ref{fig:diff_sizes_error}. 
The results show that 80k preference pairs lead to strong performance, surpassing even the 100k case. 
Based on this finding, we select 80k preference pairs to optimize the preference-specific reward model.
Similar conclusions are drawn when testing for the optimal number of selected preference pairs under the same settings on the Mistral model.

\begin{figure}[t]
    \centering
    \vspace{1.5mm}
    \includegraphics[width=1.0\linewidth]{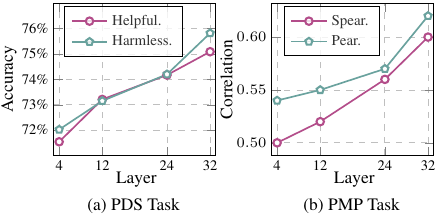} 
    \vspace{-7mm}
    \caption{
    Performance of preference representations at different layers on the PDS and PMP tasks.
    }
    \vspace{-5.5mm}
    \label{fig:pds_pmp_diff_layers}
\end{figure}

\paragraph{Probing Preference Representations at Different Layers.} 
We probe preference representations at different layers of the representation model.
Specifically, we visualize the learned preference representations of unseen preference pairs on different preference types and tasks, at layers 4, 12, 24, and 32 in Figure \ref{fig:different_layers}.
We also present an evaluation of these representations on the PDS and PMP tasks in Figure \ref{fig:pds_pmp_diff_layers}.
The visualizations reveal that the learned representations successfully distinguish samples based on the preference features they encompass. 
However, at shallower layers, the model captures preference features like preference type less effectively in a representation, with improved representation as the number of layers increases. 
In contrast, the task feature is captured more effectively at shallow layers, likely due to their relative simplicity compared to preference type.
Consistent with the visualizations, both PDS and PMP performance improves as the number of layers increases.
Interestingly, we see that although the visual distinction between the 24-layer and 32-layer representations is minimal, the 32-layer representation outperforms the 24-layer one on the PDS and PMP tasks.
We conjecture that between layers 24 and 32, the model may have learned or refined additional features relevant to human preferences, which are useful for improving the performance of these tasks.

\paragraph{Broader Applications of Preference Representation.}
We discuss potential broader applications of preference representation: 

(1) Preference data filtering.
Preference data filtering can significantly improve DPO training \cite{morimura2024filtered}. 
Future work could leverage preference representations for filtering, where we label a small subset of data for quality evaluation and fine-tune the representation model via adaptive learning. 
The model can then be applied to evaluate the quality of preference data at scale.

(2) Preference data retrieval.
Building upon the approach for demonstration and preference retrieval \cite{chang2024efficient,anonymous2024retrievaldpo}, we could retrieve preference pairs for prompt construction or dataset creation through preference representations learned by LRHP.

Beyond that, we believe there are broader applications for preference representation.
Our work only makes the very first attempt, and we are hoping this work can inspire further research on preference representation learning.

\section{Conclusion}
In this paper, we introduced the preference representation learning task and proposed LRHP to tackle it.
We further investigated the use of preference representations through two designed downstream tasks: preference data selection and preference margin prediction. 
The experimental results show that LRHP can effectively learn richer preference representations of human preferences.

\section*{Limitations}
First, we did not test models larger than 8B. However, we believe that our results using the LLaMA-3-8B-Instruction and Mistral-7B-Instruction models are sufficient for practical applications, particularly in scenarios with limited computational resources. 
Moreover, the LLaMA-3-8B-Instruction model has demonstrated superior performance compared to many larger LLMs.
Second, we primarily explore the preference representation learning on preference pairs, though individual responses (\textit{i.e.}, demonstrations) are also associated with human preferences \cite{sun2024inverse}. 
Future work could investigate how to learn preference representations directly from demonstrations like inverse reinforcement learning \cite{arora2021survey}.
Finally, our preference representation does not directly improve human preference alignment but requires training through reward modeling and DPO. 
Future work could explore using this representation to directly improve alignment, such as by incorporating it into an LLM to refine its output.

\section*{Ethics Statement}
No ethical considerations are required for this work. 
Although preference margin scores are labeled as described in Section \ref{sec:task2}, the preference pairs are derived from open-source data, and all annotators have been uniformly trained for the labeling process.
Furthermore, we will publicly release the labeled data upon the paper's publication.

\section*{Acknowledgments}
This work was supported in part by the National Science Foundation of China (No.62276056), the Natural Science Foundation of Liaoning Province of China (2022-KF-16-01), the Fundamental Research Funds for the Central Universities (Nos. N2216016 and N2316002), the Yunnan Fundamental Research Projects (No. 202401BC070021), and the Program of Introducing Talents of Discipline to Universities, Plan 111 (No.B16009).

\bibliography{custom}

\appendix

\clearpage

\section{Experimental Details}
\label{app:experimental_details}
\subsection{Settings}

\paragraph{Preference Representation Model Training.}
The preference pairs used to optimize our preference representation model are presented in Table \ref{tab:fusion_preference_pairs}.
To preserve the generalization of the model, we did not utilize the entire dataset for each task-specific preference dataset. 
Instead, we randomly selected 20k pairs from the larger task-specific preference datasets.
We finally selected over 849k preference pairs to optimize our model.
The model was trained on these pairs for one epoch with a learning rate of 1e-5 and a batch size of 128, completed within over 36 hours using eight A800 GPUs.
For the PMP task, we fine-tuned the proposed preference representation model for 3 epochs using a learning rate of 2e-5 and a batch size of 32.

\paragraph{Reward Model Training.}
We trained a reward model from the preference pairs following \citet{ouyang2022training}, where the training objective was based on the Bradley-Terry model \cite{bradley1952rank}.
In the PDS task, we initialized the reward model using the LLaMA-3-8B-Instruction and Mistral-7B-Instruction models, respectively.
It is important to highlight a deviation from the conventional practice of fine-tuning language models, which often involves a reduction of the learning rate during the fine-tuning stage, as suggested by \cite{devlin2018bert,wang2023improved}. 
We opted not to decrease the learning rate during the fine-tuning of the reward model with preference-specific data pairs. 
This decision was made because the preference-specific preference data was typically very small, and we did not update the trained pre-trained reward model many times.

\paragraph{Best-of-$n$ Sampling.}
When conducting the best-of-$n$ sampling on the PDS task, we generated eight candidate responses using the top-$p$ sampling approach, where we set $p$ to 0.95 and temperature to 0.75.
Then, we picked a final generated output that had the maximum reward score.

\paragraph{RL Training.}
We trained the LLM using PPO via the \texttt{trlx} implementation\footnote{\url{https://github.com/CarperAI/trlx}}. 
For this training, the Anthropic Helpful\&Harmless dataset was utilized to evaluate the performance of the PDS task.
For all experiments, the learning rate was set to 1e-5 and 5e-6 for the policy model and the value model, respectively.
We settled on a batch size of 64 for each PPO step, which consisted of 1 epoch of gradient steps and 4 epochs of mini-batch PPO steps.
To address the over-optimization issue as described in \citet{gao2023scaling}'s work, we implemented a strategy that saved checkpoints at regular intervals during the training process.
Specifically, we evaluated checkpoints at intervals of 200 steps for all tasks against their respective validation sets and selected the optimal checkpoint with the best Reward score.
Following \citet{wang2024hybrid}, we also employed a cold-start trick for PPO, to alleviate the damage caused by the inaccurate estimation of the early value model.
Specifically, we only updated the value model and did not update the policy model during the first 50 steps of PPO training.
Additionally, following \citet{wang2024esrl}'s work, we standardized our reward scores using a reward queue, which stored the previous 1k reward scores to calculate the mean and variance.
All of our experiments were done on eight A800 GPUs.

\begin{table}[t]
    \centering
    \scalebox{0.72}{
    \begin{tabular}{lcrc}
\toprule[1.1pt]
Dataset                  & \multicolumn{1}{l}{Scoure} & Num. & Use \\  \midrule
\textbf{\textit{General Preference}}       & \multicolumn{1}{l}{}       & \multicolumn{1}{l}{}                                                    &      \\ \midrule
Anthropic \cite{bai2022training}                & Human                      & 86k                                                                     & All  \\
Alpacafarm \cite{dubois2024alpacafarm}               & Human                      & 9k                                                                      & All  \\
SHP \cite{ethayarajh2024kto}                      & Human                      & 348k                                                                    & All  \\
UltraFeedback \cite{cui2023ultrafeedback}            & AI                         & 321k                                                                    & All  \\ \midrule
\textbf{\textit{Task-Specific Preference}} & \multicolumn{1}{l}{}       &                                                                         &      \\ \midrule
WebQA \cite{nakano2021webgpt}                    & Human                      & 14k                                                                     & All  \\
Summarization \cite{stiennon2020learning}            & Human                      & 92k                                                                     & 20k  \\
Math Reasoning \cite{lai2024step}           & Rule                       & 10k                                                                     & All  \\
Code Generation \cite{CodeFeedback-Filtered-Instruction}           & Rule                       & 54k                                                                     & 20k  \\
Machine Translation \cite{xu2024contrastive}      & COMET                      & 24k                                                                     & 20k  \\ \midrule
\end{tabular}}
    \caption{
    We provide the labeling approach for each dataset in the “Source” column, the number of preference pairs in the “Num.” column, and the number of datasets we use per dataset in the “Use” column.
    }
    \vspace{-4mm}
    \label{tab:fusion_preference_pairs}
\end{table}

\begin{figure*}[t]
    \centering
    \includegraphics[width=1.0\textwidth]{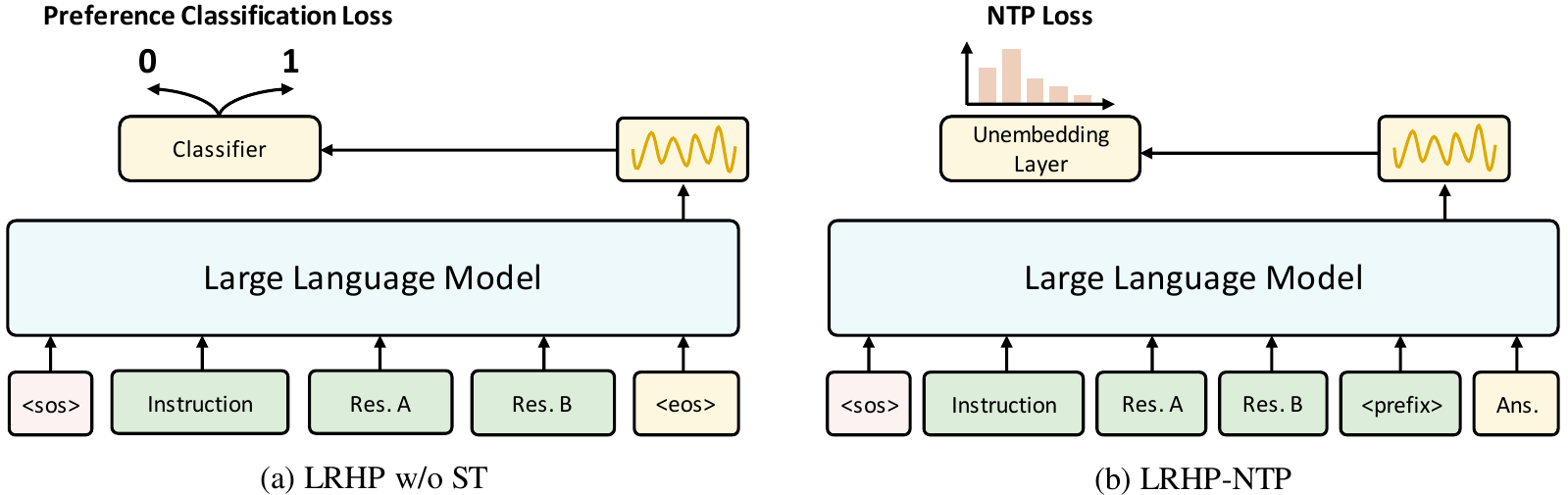}
    \vspace{-7mm}
    \caption{
    The architectures of LRHP w/o ST and LRHP-NTP.
    In LRHP-NTP, we use the hidden output from the answer token prediction as the preference representation.
    }
    \vspace{-4mm}
    \label{fig:other_optimization_approaches}
\end{figure*}

\subsection{Evaluation}
We explain the approach used to calculate the win rate in the PDS task experiments, as follows.
Given the pairwise test responses $\{(x^{0}, r_a^{0},  r_b^{0}), \cdots, (x^{T}, \\ r_a^{T}, r_b^{T})\}$, where $T$ is the number of the test set, we employed GPT-4 to annotate the preference of each pairwise response, including $P_a$, $P_b$, and $Tie$.
Here, $P_a$ denotes response $r_{a}$ is better than response $r_{b}$, $P_b$ denotes response $r_{b}$ is worse than response $r_{b}$, while $Tie$ denotes a tie between response $r_{a}$ and response $r_{b}$.
To address potential location bias in the evaluation \cite{gao2024llm}, we conducted two separate evaluations for each pair, alternating the order of $r_a$ and $r_b$. 
The final result is based on the evaluations where the preferences align consistently.
We computed the win rate for the models generating responses $r_a$ and $r_b$ based on these annotated preferences:
\begin{eqnarray}
    S_{\mathrm{WinRate}}^{a}=\frac{\mathrm{Count}(P_{a})}{T-\mathrm{Count}(Tie)} \\
    S_{\mathrm{WinRate}}^{b}=\frac{\mathrm{Count}(P_{b})}{T-\mathrm{Count}(Tie)}
\end{eqnarray}
where $\mathrm{Count}(\cdot)$ represents the number of occurrences of the specified preference.
For the computation of win rates in the PMP task, we utilized auto-evaluation systems including alpaca\_eval\footnote{\url{https://github.com/tatsu-lab/alpaca_eval}} and arena-hard-auto\footnote{\url{https://github.com/lm-sys/arena-hard-auto}}.

\begin{table}[ht]
    \centering
    \scalebox{0.70}{
    \begin{tabular}{lcccc}
\toprule[1.1pt]
             & \begin{tabular}[c]{@{}c@{}}Negligibly\\ Better / Unsure\end{tabular} &  \begin{tabular}[c]{@{}c@{}}Slightly\\ Better\end{tabular} &
             Better
             &
             \begin{tabular}[c]{@{}c@{}}Significantly\\ Better\end{tabular}   \\  \midrule
Margin Score & 1                                                                    & 2                                                              & 3      & 4       \\ \bottomrule[1.1pt]                                                 
\end{tabular}}
    \caption{
    We draw upon the work of \citet{touvron2023llama} to establish our criteria for labeling preference margins.
    The score represents the degree of difference by which the preferred response surpasses the dispreferred one within a preference pair.
    }
    \vspace{-4mm}
    \label{tab:margin_labeling}
\end{table}

\subsection{Preference Margin Labeling}
We employed two undergraduate and two graduate students as our annotators. 
Before labeling, we conducted training for them to standardize the labeling criteria. 
Specifically, we assigned a score of 1, 2, 3, or 4 to each preference pair to serve as the corresponding margin score. 
The specific meanings of these scores are indicated in Table \ref{tab:margin_labeling}. 
These scores describe the overall difference between the comparison pairs, rather than assessing any specific aspect.
After labeling, we converted these scores into real values and applied min-max normalization to facilitate the optimization of the preference representation model.

Upon publication of this paper, we will make our preference representation model, labeled preference margin data, and the accompanying source code available as open-source.

\section{More Analysis}
\label{app:more_analysis}

\begin{figure}[t]
    \centering
    \vspace{1mm}
    \includegraphics[width=0.48\textwidth]{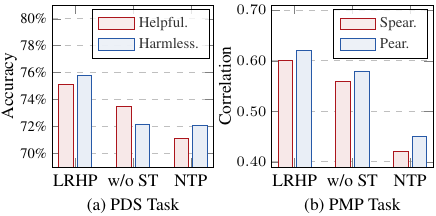}
    \vspace{-6mm}
    \caption{
    We compare alternative optimization approaches for the preference representation model, including LRHP without a specific token (namely w/o ST) and LRHP-NTP.
    }
    \vspace{-5mm}
    \label{fig:diff_optimization_approaches}
\end{figure}

\paragraph{Different Optimization Approaches for Preference Representation Model.}
In addition to preference classification, we explore alternative optimization approaches: LRHP without a specific token (w/o ST) and LRHP with the next-token prediction (LRHP-NTP).
In LRHP w/o ST, we modify the original LRHP by removing the \texttt{<|PREFERENCE|>} token and replacing it with \texttt{<eos>}. 
In LRHP-NTP, we adopt a standard language modeling task instead of preference classification to optimize the preference representation model. 
The architectures of these optimization approaches are illustrated in Figure \ref{fig:other_optimization_approaches}. 
We compare their performance on the PDS and PMP tasks in Figure \ref{fig:diff_optimization_approaches}.
The results show that LRHP achieves the best performance, likely because using \texttt{<eos>} or NTP tokens as preference representations is suboptimal. 
These tokens are pre-trained and may carry meanings that interfere with the effective learning of preference representations.

\begin{figure*}[t]
    \includegraphics[width=1.0\textwidth]{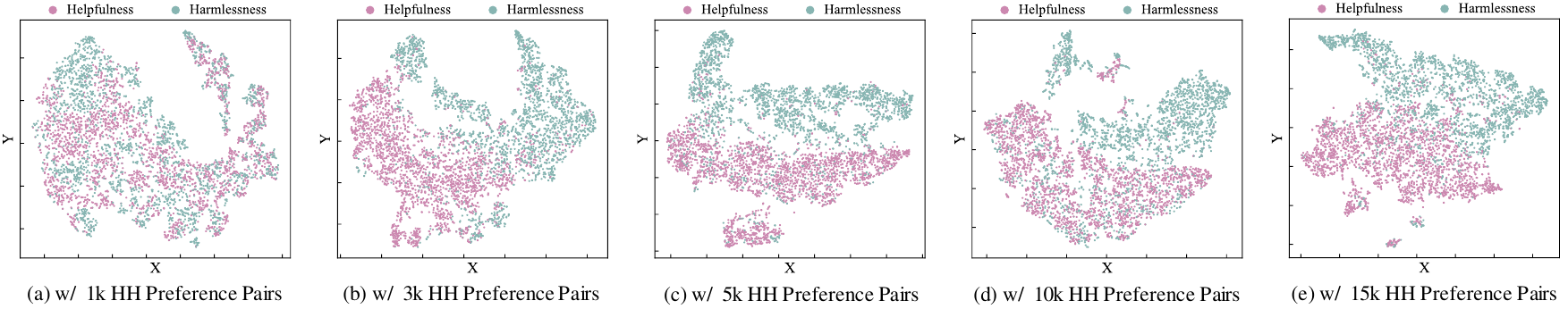}
    \vspace{-7mm}
    \caption{
    We visualize the learned preference representations of the model optimized using different numbers of Anthropic Helpful\&Harmless preference pairs.
    }
    \label{fig:vis_diff_number_ps_pairs}
\end{figure*}

\begin{figure}[t]
    \centering
    \vspace{-4mm}
    \includegraphics[width=0.48\textwidth]{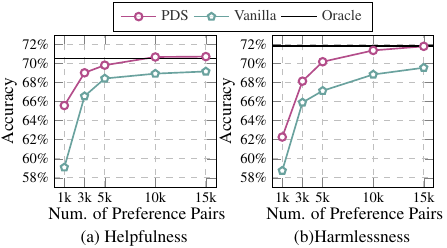}
    \vspace{-6mm}
    \caption{
    We evaluate the PDS task performance using different numbers of Anthropic Helpful\&Harmless preference pairs: 1k, 3k, 5k, 10k, and 15k. 
    “Oracle” denotes the reward model trained on 86k preference pairs.
    }
    \vspace{-4mm}
    \label{fig:pds_diff_number_ps_pairs}
\end{figure}

\paragraph{Enabling Few-Shot Learning in Reward Model.}
Figure \ref{fig:vis_diff_number_ps_pairs} presents a visualization of our preference representation model, optimized using different numbers of Anthropic Helpful\&Harmless data to simulate a few-shot learning scenario. 
Note that the numbers for the remaining datasets, as listed in Table \ref{tab:fusion_preference_pairs}, remain unchanged for optimization.
The results indicate that with limited preference-specific data, LRHP successfully learns representations that capture corresponding human preferences. 
Additionally, as shown in Figure \ref{fig:pds_diff_number_ps_pairs}, we evaluate performance on the PDS task. 
Pre-training with selected preference data facilitates effective few-shot learning in the reward model, enabling it to quickly generalize to unseen data with a few preference-specific preference pairs. 
Notably, using only 5k preference-specific pairs achieves performance comparable to that of the reward model trained with 86k preference pairs.

\end{document}